\documentclass{article}

     \usepackage[preprint, nonatbib]{neurips_2021}

\usepackage[utf8]{inputenc} 
\usepackage[T1]{fontenc}    
\usepackage{hyperref}       
\usepackage{url}            
\usepackage{booktabs}       
\usepackage{amsfonts}       
\usepackage{nicefrac}       
\usepackage{microtype}      

\usepackage{graphicx}
\usepackage{subfigure}
\usepackage{bm}
\usepackage{amsmath}
\usepackage{amssymb}
\usepackage{multicol}
\usepackage{multirow}
\usepackage{enumitem}
\usepackage{color}
\usepackage{wrapfig,lipsum,booktabs}
\usepackage{bbm}

\title{Adaptive Label Smoothing To Regularize \\ Large-Scale Graph Training}

\author{%
  Kaixiong Zhou \\
  Rice University\\
  \texttt{Kaixiong.Zhou@rice.edu} \\
   \And
   Ninghao Liu \\
   University of Georgia \\
   \texttt{ninghao.liu@uga.edu} \\
   \And
   Fan Yang \\
   Rice University \\
   \texttt{fyang@rice.edu} \\
   \And
   Zirui Liu \\
   Rice University \\
   \texttt{Zirui.Liu@rice.edu}
   \\
   \And
   Rui Chen \\
   Samsung Research America \\
   \texttt{rui.chen1@samsung.com} \\
   \And
   Li Li \\
   Samsung Research America \\
   \texttt{li.li1@samsung.com} \\
   \\
   \And
   Soo-Hyun Choi \\ 
   Samsung Electronics \\ 
   \texttt{soohyunc@gmail.com} \\
    \\
   \And
   Xia Hu \\
   Rice University \\
   \texttt{xia.hu@rice.edu}
}

\begin{document}

\maketitle

\begin{abstract}
Graph neural networks (GNNs), which learn the node representations by recursively aggregating information from its neighbors, have become a predominant computational tool in many domains. To handle large-scale graphs, most of the existing methods partition the input graph into multiple sub-graphs (e.g., through node clustering) and apply batch training to save memory cost. However, such batch training will lead to label bias within each batch, and then result in over-confidence in model predictions. Since the connected nodes with positively related labels tend to be assigned together, the traditional cross-entropy minimization process will attend on the predictions of biased classes in the batch, and may intensify the overfitting issue. To overcome the label bias problem, we propose the adaptive label smoothing (ALS) method to replace the one-hot hard labels with smoothed ones, which learns to allocate label confidences from the biased classes to the others. Specifically, ALS propagates node labels to aggregate the neighborhood label distribution in a pre-processing step, and then updates the optimal smoothed labels online to adapt to specific graph structure. Experiments on the real-world datasets demonstrate that ALS can be generally applied to the main scalable learning frameworks to calibrate the biased labels and improve generalization performances. 
\end{abstract}

\section{Introduction}
Large-scale graphs, which are characterized by massive nodes and edges, are ubiquitous in real-world applications, such as social networks~\cite{ching2015one, wang2018billion, perozzi2014deepwalk} and knowledge graphs~\cite{wang2018acekg, wang2018acekg}. 
Although graph neural networks (GNNs) have shown effectiveness in many fields~\cite{kipf2016semi, velivckovic2017graph, wu2019simplifying, chen2020simple, xu2018powerful}, most of them rely on propagating messages over the whole graph dataset, and are mainly developed for relatively small graphs. Such message passing paradigms lead to prohibitive computation and memory requirements. 

Recently, several scalable algorithms of GNNs have been proposed to handle the large-scale graphs, among which sub-graph sampling methods are dominant in literature~\cite{hamilton2017inductive, chen2018fastgcn,  chen2017stochastic, chiang2019cluster, zeng2019graphsaint}. Specifically, instead of training on the full graph, the sampling methods sample subsets of nodes and edges to formulate a sub-graph at each step, which is treated as an independent mini-batch. For example, Cluster-GCN~\cite{chiang2019cluster} first clusters the input graph into sub-graph groups, and then formulates each batch with a fixed number of groups (referred as batch size) during model training. LGCN~\cite{gao2018large} samples sub-graphs via breadth first search,
as motivated by small patch cropping on a large image.  

\begin{figure}[ht]
    \centering
    \includegraphics[width=\textwidth]{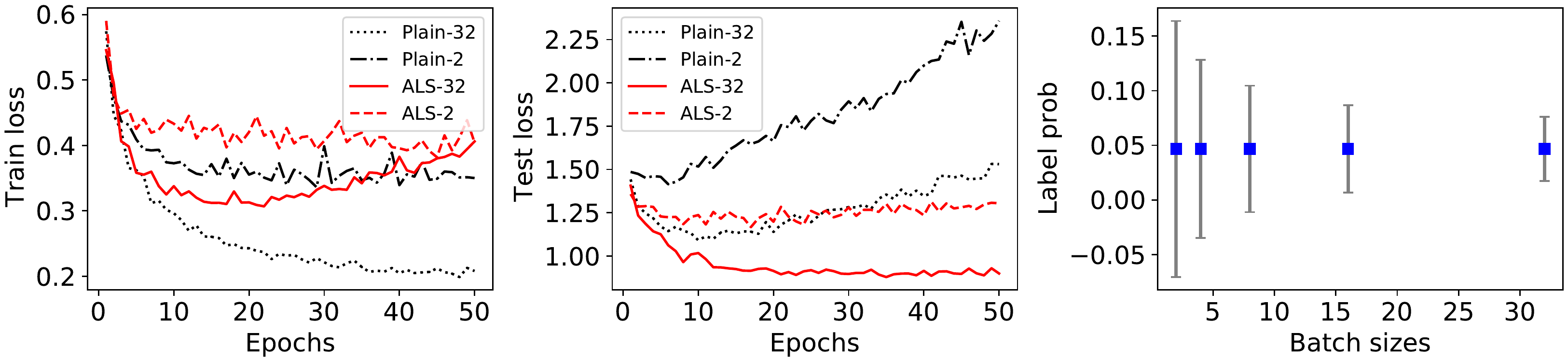}
    \caption{\textbf{Left, Middle:} The training and testing losses on ogbn-products. While Plain-$a$ means the original Cluster-GCN trained with batch size of $a$, ALS-$a$ denotes the Cluster-GCN equipped with ALS. \textbf{Right:} The mean probability $p_c$ of nodes with specific class $c$ within a batch, and the standard variance of probability $p_c$ among batches. Herein we show $c=0$ on  ogbn-products for an example.}
    \label{fig:loss}
\end{figure}

Nevertheless, the label bias existing in the sampled sub-graphs could make GNN models become over-confident about their predictions, which leads to overfitting and lowers the generalization accuracy~\cite{ghoshal2020learning}. Note that in the real-world assortative graphs~\cite{tang2013exploiting}, the closely connected nodes are potential to share the same label or positively related labels. The sub-graph sampling methods usually assign these related nodes into the same sub-graph and lead to label bias in a batch, whose node label distribution is significantly different from the other batches. Taking Cluster-GCN as an example, where the community with the same node labels is clustered as a sub-graph, the label distribution variance among batches is dramatic as shown in Figure~\ref{fig:loss}. Comparing with the traditional deep neural networks trained by uniform batch~\cite{muller2019does, li2020regularization, ding2019adaptive, xu2020towards}, the cross-entropy minimization in the biased batch will severely make GNN model to attend only on the correctness of biased ground-truth category by producing an extremely large prediction probability. Such over-confident prediction will overfit on training set (e.g., the decreasing training loss of Cluster-GCN in Figure~\ref{fig:loss}), but generalizes poorly on the testing set (e.g., the increasing testing loss).  To overcome the over-prediction and overfitting, label smoothing has been proposed to soften the one-hot class label by mixing it with a uniform distribution~\cite{szegedy2016rethinking}. Through penalizing the over-confident prediction towards the ground-truth label, the label smoothing (LS) has been used to improve the generalization performance across a range of tasks, including image classification~\cite{muller2019does, xu2020towards, lienen2021label, yuan2020revisiting}, semantic parsing~\cite{ghoshal2020learning}, and neural machine translation~\cite{meister2020generalized, gao2020towards}.

However, it is non-trivial to apply the label smoothing to regularize and adapt to the large-scale graph training from two structural levels: local node and global graph. First, different from generic machine learning problems associated with independent data instances, in graph data, it is generally assumed that class labels of connected nodes are positively related in many real-world applications~\cite{tang2013exploiting, huang2020combining}. In other words, for a specific local node, its label prediction highly depends on the label distribution of its neighbors. The traditional label smoothing, after mixing one-hot hard target with a uniform distribution, could wrongly regularize nodes in graph data. Second, considering the global graph, the relevance between different pairs of labels could vary. For example, in academic networks, civil engineering researchers tend to collaborate more with ecology researchers than with physicists~\cite{newman2004coauthorship}. The optimal label smoothing should be conditioned on the the relevance between the ground-truth label and the related labels. The fixed label smoothing paradigm, by mixing uniform distribution, will fail to model such global label relevance specifically to the downstream tasks. 

To bridge the gap, in this paper, we develop a simple yet effective adaptive label smoothing (ALS) method to regularize the representation learning on large-scale graphs. We aim to answer two research questions. First, given a local node, how can we estimate its smoothed label that is aware of the neighborhood structure? Second, how can we learn the global label relevance for a specific task? Through exploring these questions, we make three significant contributions as follows.
\begin{itemize}[leftmargin=*, topsep=0pt, noitemsep]
    \item We are the first to analyze the label bias problem in the sub-graph sampling methods for the large-scale graph training. The biased batch training could make GNN model produce over-confident prediction and overfit on the training set.
    
    \item We present an adaptive label smoothing methods decoupled into the following stages: a label propagation preprocessing step to aggregate the local neighborhood label distribution; a label refinement step mapping the preprocessed neighborhood labels to learn the global smoothed label adpatively. Our method is very simple and memory efficient, and could be scaled to the large-scale graph with negligible step to map the desired smoothed label.
    
    \item We propose a label smoothing pacing function to allocate different smoothing strengths along the training process, in order to avoid the overly regularization at the beginning.
    \item The empirical results show that our adaptive label smoothing could relieve the overfitting issue and yield better node classification accuracies based upon most scalable learning frameworks. 
\end{itemize}

\section{Label Bias, Over-confident Prediction and Overfitting}
\paragraph{Notations and problem definition.} We denote matrices with boldface capital letters (e.g. $\bm{A}$), vectors with boldface lowercase letters~(e.g., $\bm{y}$ or $\bm{y}_i$) and scalars with lowercase alphabets~(e.g., $a$). We use $A_{i,j}$ to index the $(i,j)$-th element in matrix $\bm{A}$, and use $y_{i,j}$ to represent the $j$-th entity of vector $\bm{y}_i$. In this work, we focus on node classification tasks, and propose using label smoothing to address the over-confident prediction and overfitting issue in large-scale graph analysis. A graph is represented by $G = (\bm{A}, \bm{X})$, where $\bm{A} \in \mathbb{R}^{N\times N}$ denotes the adjacency matrix, $\bm{X}\in \mathbb{R}^{N\times d}$ denotes the feature matrix, and $N$ is the number of nodes. Each node $i \in \mathcal{V}$ is associated with a feature vector $\bm{x}_i \in \mathbb{R}^{d}$ (indexed by the $i$-th row in $\bm{X}$) and a one-hot class label $\bm{y}_i \in \mathbb{R}^{C}$, where $C$ is the number of class labels. Given a training set $\mathcal{V}_l$ with labels, the goal is to classify the nodes in the unlabeled set $\mathcal{V}_u = \mathcal{V}\setminus \mathcal{V}_l$ via learning effective node representations. Let $f_{\theta}$ denote the GNN model, where $\theta$ denotes model parameters. The prediction for a node is $\hat{\bm{y}}_i = f_{\theta}((\bm{A}, \bm{X})) \in \mathbb{R}^{C}$.
Recalling the batch training in a large-scale graph, the plain cross-entropy loss in a batch $\mathcal{B}$ is given by: 
\begin{equation}
\label{eq:plain_loss}
   \mathcal{L}^{\mathrm{Plain}}(\theta) = \frac{1}{|\mathcal{B}|}\sum_{i\in \mathcal{B}}H(\bm{y}_i, \hat{\bm{y}}_i) = -\frac{1}{|\mathcal{B}|}\sum_{i\in \mathcal{B}}\sum_{c=1}^{C}y_{i,c}\log{\hat{y}_{i,c}}, 
\end{equation}
where $H$ is the cross-entropy function. Sub-graphs are sampled to build each batch, and $\mathcal{B}$ contains the nodes in the sampled sub-graphs for training. $|\mathcal{B}|$ denotes the number of training nodes in $\mathcal{B}$.


\paragraph{Model analysis.} We empirically study the label bias phenomenon, over-prediction and overfitting issues of Cluster-GCN models trained on the ogbn-products dataset. Other sub-graph sampling methods with the similar issue are shown in Appendix. First, to evaluate the label bias, we define the probability of nodes with class $c$ in a batch as: $p_c=\frac{\sum_{i\in \mathcal{B}}\mathbbm{1}_{{y}_{i,c}=1}}{|\mathcal{B}|}$, where $\mathbbm{1}_{{y}_{i,c}=1}$ indicates whether node $i$ belongs to class $c$. The mean and standard variance of $p_c$ among batches are shown in the right part of Figure~\ref{fig:loss}. We could observe that the standard variance is significantly larger when batch size is small, while the mean value is relatively stable. It means that nodes within a small batch tend to belong to certain classes, instead of evenly distribute across all classes. Such label bias is inherent in sub-graph sampling methods, such as clustering~\cite{chiang2019cluster} and random walk sampling~\cite{zeng2019graphsaint}, since positively related nodes are more likely to be selected together in a sampled sub-graph.

We further analyze the training and testing losses of Cluster-GCN in Figure~\ref{fig:loss}. Different from using uniform batch labels in traditional machine learning, the label bias in the batches further leads the GNN model to over-confidently attend (i.e., produce large prediction probabilities) on the ground-truth classes by minimizing the vanilla cross-entropy loss. The over-confident prediction overfits the training set and accelerates the decrease of training loss, but poorly generalizes to the testing set as shown in the increased testing loss. Comparing with the batch size of $2$, a larger batch size of $32$ reduces the variance of $p_c$ and relieves the label bias to some extent, which brings a smaller testing loss and better generalization performance. However, the big batch size would improve computation and memory costs, which is not inline with the purpose of sub-graph batch training on large graphs.

\paragraph{Label smoothing.} 
To overcome over-confidence and overfitting, label smoothing has been proposed to mix the one-hot hard labels with uniform distribution in the image classification~\cite{muller2019does, xu2020towards, lienen2021label, yuan2020revisiting} and natural language processing~\cite{ghoshal2020learning, meister2020generalized}. 
To be specific, considering a training node $i$, its smoothed label is given by: $\bm{y}_i^{LS} = (1-\alpha)\bm{y}_i + \alpha \bm{1}/C$, where $\bm{1}/C = [1/C, \dots, 1/C]^{\top} \in \mathbb{R}^C$ is a uniform distribution and $\alpha$ is the regularization strength. Then the cross-entropy is given by:
\begin{equation}
\label{eq: ls}
\mathcal{L}^{LS}(\theta) = \frac{1}{|\mathcal{B}|}\sum_{i\in \mathcal{B}} H(\bm{y}_i^{LS}, \hat{\bm{y}}_i) = \frac{1}{|\mathcal{B}|}\sum_{i\in \mathcal{B}}(1-\alpha) H(\bm{y}_i, \hat{\bm{y}}_i) + \alpha H(\bm{1}/C, \hat{\bm{y}}_i).
\end{equation}
By minimizing $\mathcal{L}^{LS}(\theta)$, we are able to prevent the model from being over-confident on the ground-truth $\bm{y}_i$, via a penalty that enforces non-nigligible prediction probabilities on the other classes. 

\section{Adaptive Label Smoothing}
The above label smoothing fails to adapt to graph data by ignoring two informative attributes: the label distribution within local neighborhood and the global label relevance. 
First, the nodes' labels are not independently distributed, and correlate positively to their local neighbors~\cite{huang2020combining, wang2020unifying, jia2020residual, zhou2004learning, zhu2003semi}. The vanilla label smoothing assumes that node labels are independent and identically distributed, and applies the uniform distribution to regularize representation learning. It could mislead the model prediction to attend on the negatively related labels. Second, in the overall graph, the relevance between each pair of labels is different from each other. For example, the pair-wise collaboration strengths among engineering, ecology, and physical researchers are unbalanced in the academic networks~\cite{newman2004coauthorship}. In the hierarchical GNNs, it is commonly assumed that the connections between labeled communities should be sparse~\cite{ying2018hierarchical, zhou2019multi}, instead of the uniform and full connection. The vanilla label smoothing with fixed uniform distribution cannot properly learn the latent global label relevance for the downstream task.  


\subsection{Proposed Techniques}
In this work, we propose ALS to calibrate the label bias and regularize the sub-graph batch training for the large-scale graph. 
ALS consists of three parts: 
(1) a pre-processing step of label propagation to obtain the prior knowledge of neighborhood label distribution; 
(2) a label refinement step to correct the prior knowledge and learn the global label relevance in the training phase; 
and (3) a smooth pacing function to gradually schedule the smooth strength and avoid the overly label regularization. 

\paragraph{Label propagation.}
Based on the expectation that the two connected nodes are likely to have the same label according to graph homophily, label propagation passes labels iteratively to learn the label predictions~\cite{jia2020residual, wang2020unifying, ando2007learning, belkin2006manifold}. 
However, most of them involve parameters and are expensive to be trained. 
To scale to large-scale graphs, we simplify the label propagation by removing the trainable weights, and conduct it only as a pre-processing step. Specifically, let $\bm{Y}^{(k)} = [\bm{y}^{(k)}_1, \cdots, \bm{y}^{(k)}_N]^\top \in \mathbb{R}^{N\times C}$ denote the propagated label matrix obtained from the $k$-th iteration of label propagation. At the $k+1$-th iteration, we update the propagated label matrix as follows:
\begin{equation}
    \label{eq: lp}
    \bm{Y}^{(k+1)} = (1-\beta)\bm{D}^{-1}\bm{A}\bm{Y}^{(k)} + \beta \bm{Y}^{(0)}.
\end{equation}
For the initial label matrix $\bm{Y}^{(0)} = [\bm{y}^{(0)}_1, \cdots, \bm{y}^{(0)}_N]^\top$, it consists of one-hot hard labels $\bm{y}_i$ for training nodes and zero vectors otherwise. $\beta$ is residual strength to preserve the initial training labels, and $\bm{D}$ is the diagonal degree matrix of $\bm{A}$. The label propagation in Eq.~\eqref{eq: lp} is similar in spirit to~\cite{zhu2002learning}, but we preserve $\bm{Y}^{(0)}$ to avoid the overwhelming of training labels. 
After $K$ iterations of label propagation, we obtain the prior knowledge of neighborhood label distribution $\bm{Y}^{(K)}$ up to $K$ hops away. Such prior knowledge provides enough neighborhood information to be refined. Our label propagation has a good trade-off between the efficiency and effectiveness. 

\paragraph{Label refinement.} 
In this step, we aim to refine propagated label matrix $\bm{Y}^{(K)}$. Specifically, given the propagated label $\bm{y}^{(K)}_i\in \mathbb{R}^{C}$ of training node $i$ (i.e., indexed from the $i$-th row of $\bm{Y}^{(K)}$), we correct it by: $\bm{y}_i^{\mathrm{soft}} = \mathrm{Softmax}(\bm{W}\bm{y}^{(K)}_i)$. $\bm{W}\in \mathbb{R}^{C\times C}$ is a trainable matrix.  The smoothed label used to regularize model training is then given by:
\begin{equation}
    \label{eq:adpative_label}
    \bm{y}_i^{ALS} = (1-\alpha)\bm{y}_i + \alpha \bm{y}_i^{\mathrm{soft}}.
\end{equation}
Notably, element $W_{c,j}$ indicates the latent relevance between classes $c$ and $j$, and is shared globally by all nodes over the graph. Considering node $i$, the real relevance to class $c$ is corrected to be proportional to $\sum_{j}W_{c,j}{y}^{(K)}_{i,j}$. To well learn the global label relevance in $\bm{W}$, we jointly train with classification task and compute the batch loss as follows: 
\begin{equation}
\label{eq: ALS_loss}
\begin{array}{rl}
   \mathcal{L}^{ALS}(\theta, \bm{W})  & =  \frac{1}{|\mathcal{B}|}\sum_{i\in \mathcal{B}} H(\bm{y}_i^{ALS}, \hat{\bm{y}}_i) + \gamma \mathrm{KL}(\bm{y}_i^{\mathrm{soft}}, \bm{1}/C) \\
     & = \frac{1}{|\mathcal{B}|}\sum_{i\in \mathcal{B}}(1-\alpha) H(\bm{y}_i, \hat{\bm{y}}_i) + \alpha H(\bm{y}_i^{\mathrm{soft}}, \hat{\bm{y}}_i) + \gamma \mathrm{KL}(\bm{y}_i^{\mathrm{soft}}, \bm{1}/C),
\end{array}
\end{equation}
where $\mathrm{KL}$ denotes the KL distance of two probability distribution vectors, and $\gamma$ is a positive hyperparameter. Compared with the traditional label smoothing in Eq.~\eqref{eq: ls}, ALS relaxes the uniform distribution to learn the optimal soft label $\bm{y}_i^{\mathrm{soft}}$ and adapt to the downstream task. On one hand, parameter $\bm{W}$ is updated to learn the global label relevance and produce a reasonable $\bm{y}_i^{\mathrm{soft}}$. 
On the other hand, the KL distance constraint is exploited to avoid $\bm{y}_i^{\mathrm{soft}}$ collapsing into the one-hot hard target $\bm{y}_i$ and guarantee the divergence. 

\paragraph{Smooth pacing.} Considering the batch training in large-scale graph, the constant smoothing strength $\alpha$ may overly regularize model at the initial training phase. Given the randomly initialized parameter $\bm{W}$, the soft label $\bm{y}_i^{\mathrm{soft}}$ will mislead model prediction $\hat{\bm{y}}_i$ to attend on the unrelated classes. Motivated from the batch pacing in curriculum learning~\cite{bengio2009curriculum, jiang2015self}, we propose a smooth pacing function to gradually schedule the appropriate smoothing strength $\alpha_t$ at the $t$-th epoch. At the early phase, since the over-confident prediction has not appeared, we use a small $\alpha_t$ to let model learn the correct prediction. With the ongoing of training, we gradually improve $\alpha_t$ to regularize model. Specially, we consider the following two categories of pacing function: (1) a linear pacing function of $\alpha_t = \min(r\cdot t, \alpha_{\max})$, where $r$ is pacing rate and $\alpha_{\max}$ is the maximum smoothing strength; (2) an exponential pacing function of $\alpha_t = \min(b\cdot \mathrm{exp}(r\cdot t), \alpha_{\max})$, where $b$ is the initial smoothing strength at epoch $t=0$. In our ALS, we replace the constant smoothing strength $\alpha$ in Eq.~\eqref{eq: ALS_loss} with $\alpha_t$.

\subsection{Model Analysis}
\paragraph{Scalability analysis.} Based on the sparse matrix multiplication, the time complexity of label propagation is $\mathcal{O}(K||\bm{A}||_0C)$, where $||\bm{A}||_0$ is number of nonzeros in $\bm{A}$. Since the label propagation is conducted in pre-processing step, it could scale to the large-scale graph on CPU platforms with large memory. We thus ignore its memory complexity. The computation of label refinement mainly lines in the matrix multiplication with  $\bm{W}$. Considering any a backbone network, the extra time complexity is only $\mathcal{O}(|\mathcal{B}|C^2)$, and the extra memory complexity is only $\mathcal{O}(C^2)$. Therefore, our ALS can augment any scalable algorithms to handle the large-scale graph.

\paragraph{Generalization analysis.}
We apply ALS to augment Cluster-GCN and train on dataset ogbn-products. 
As shown in Figure~\ref{fig:loss}, comparing with the plain Cluster-GCN, our ALS has larger training losses but is accompanied with smaller testing losses. In other word, our model has better generalization performance on testing test by avoiding the overfitting on training set. Specially, the label bias problem is much severe in the small batch size of $2$. In this case, the testing loss of plain Cluster-GCN increases significantly due to the extremely over-confident prediction and the overfitting on training set. In contrast, our model could still avoid the over-confident prediction by even increasing the training loss at the end of training. The label smoothing regularization brings and maintains a lower testing loss.

\paragraph{Comparison to previous work.} Although the label smoothing has been applied in computer vision and natural language processing~\cite{muller2019does, li2020regularization}, it has not been studied to regularize GNN models for the graph data analytics. The previous GNNs are mainly developed to process small graph. In this paper, we observe the label bias problem resulted from the sub-graph batch training in the large-scale graph, and analyze the over-confident prediction and overfitting issue. Compared with the traditional label smoothing with uniform distribution, we propose ALS to adapt to the graph data. We are aware that recently there have been some label smoothing works to learn the soft label~\cite{ghoshal2020learning, ding2019adaptive}, which is similar to our label refinement module. However, they are not targeted for graph data, missing to incorporate the graph structure. In the experiments, we empirically demonstrate that all the three modules in ALS are crucial to regularize the large-scale graph training. 

Another similar line of work is label propagation. Most of previous methods involve trainable weights and cannot scale to the large-scale graph~\cite{jia2020residual, wang2020unifying, ando2007learning, belkin2006manifold}. For those simple and scalable methods, they either directly use the propagated labels to predict the testing nodes~\cite{zhu2002learning}, or concatenate them to node features as the nodes' inputs~\cite{sun2021scalable}. In this work, we exploit the pre-processed soft label to regularize the model training. In the experiments, we empirically show that the label smoothing is a better way to exploit this prior label knowledge.

\section{Experiments}
In this section, we empirically evaluate the effectiveness of ALS on several real-world datasets. Overall, we aim to answer four research questions as follows. \textbf{Q1}: Can ALS effectively regularize the model to obtain better generalization performance, comparing with the plain model and label smoothing with uniform distribution?  \textbf{Q2}: How does each module of ALS affect its performance? \textbf{Q3}: How does ALS preform comparing with the other exploitation ways of prior label knowledge? \textbf{Q4}: How do the hyperparameters influence the performance of ALS? 

\subsection{Experiment Setup}
\paragraph{Datasets.} We evaluate our proposed models on $4$ graphs with different scales using node classification tasks, following the previous large-scale graph representation learning efforts. These benchmark datasets include Flickr~\cite{zeng2019graphsaint}, Reddit~\cite{hamilton2017inductive}, ogbn-products and ogbn-mag~\cite{hu2020open}, whose node numbers are 89K, 233K, 2449K and 1940K, respectively. Their data statistics are provided in Appendix. 

\paragraph{Backbone frameworks.}
We mainly evaluate ALS on the scalable backbone frameworks based upon sub-graph sampling. Since the precomputing methods are another important lines of scalable graph representation learning, we conduct our method on them to demonstrate the general effectiveness. For the sub-graph sampling based methods, we adopt the popular backbone frameworks of GraphSAGE~\cite{hamilton2017inductive}, Cluster-GCN~\cite{chiang2019cluster} and GraphSAINT~\cite{zeng2019graphsaint}. For the pre-computing based methods, we choose the backbone frameworks of MLP and SIGN~\cite{rossi2020sign}. The detailed descriptions of these five frameworks are provided in Appendix. 
Note that we aim to demonstrate the general effectiveness of ALS in improving model generalization for the diverse scalable learning frameworks, instead of achieving the state-of-the-art performance on each classification task. Therefore, for each experiment on benchmark datasets, we conduct and compare three implementations: the plain scalable model trained with cross-entropy loss in Eq.~(\ref{eq:plain_loss}), the model augmented with conventional label smoothing (LS) as shown in Eq.~(\ref{eq: ls}), and the model augmented with ALS as shown in Eq.~(\ref{eq: ALS_loss}).

\paragraph{Implementation.} We directly use the implementations of the backbone networks either from the their official repositories or based on the official examples of PyTorch Geometric. We further implement LS and ALS over each backbone model. For LS with uniform distribution, we set the constant smoothing strength $\alpha$ as $0.1$, which is widely applied in regularizing image classification. For our ALS, we choose the appropriate hyperperameters of residual strength $\beta$ and step $K$ in the label propagation, and also determine the KL distance constraint $\gamma$ as well as the smooth pacing rate $r$. While the linear smooth
pacing is adopted in the sub-graph sampling methods, the exponential pacing function is used in the precomputing methods. The detailed choices on four datasets are shown in Appendix. We study the influences of these hyperparameters in the experiments, and show that our model is not sensitive to them within a wide value range.  

\subsection{Experiment Results}

\paragraph{Generalization improvement by label smoothing.}
To provide answers for the research question \textbf{Q1}, Table~\ref{tab: full} summarizes the comprehensive comparisons among the plain model without any label smoothing, the regularized model with LS, and the regularized model with ALS over each combination of backbone framework and dataset. It is observed that our ALS can achieve superior performances in $18$ cases out the 20 in total. Specifically, compared with the plain frameworks based on sub-graph sampling, both LS and ALS can generally improve test accuracy. The sampling methods assign connected nodes possibly with the same label into a sub-graph, which will lead to label bias within a batch. 
The label bias will make model over-confidently attend on the prediction of the ground-truth class, and may mislead model to fall into local minimums and decrease its generalization ability. While LS uses the uniform distribution to regularize model's prediction on other classes, our proposed ALS calibrates model more accurately by considering the local neighbors and using the global label refinement to correct the smoothed label.  

Compared with the plain precomputing model, our ALS can still generally improve the test accuracy, although LS tends to deteriorate model performance. Trained on GeForce RTX 2080 Ti GPU, the official implementations of MLP and SIGN use a full batch of training nodes. Since the label bias is not a big concern in such full batch training scenarios, the crude LS over-regularizes models and further hinders the accurate predictions on the ground-truth class. Notably, ALS exploits the informatic neighborhood label distribution to calibrate the model prediction, considering that the connected nodes should be close in the label space. Furthermore, the label refinement module learns to correct the smoothed label and jointly trains with the cross-entropy classification loss, which could adapt to the desired classification task. 

\begin{table}[t]
\setlength{\tabcolsep}{5.5pt}
  \centering
  \begin{tabular}{c|c|c|c|c|c}
    \toprule
    Base frameworks & Methods & Flickr & Reddit & ogbn-products & ogbn-mag\\
     \hline
     \hline
     \multirow{3}*{GraphSAGE}& Plain &  52.12$\pm$0.33 & 96.04$\pm$0.07 & 78.45$\pm$0.35 & 46.70$\pm$0.30  \\
      &LS & 52.05$\pm$0.42 & 96.23$\pm$0.08 & 78.51$\pm$0.32 & 46.87$\pm$0.55 \\
     & ALS & \textbf{52.34$\pm$0.21} & \textbf{96.26$\pm$0.08} & \textbf{78.64$\pm$0.46} & \textbf{47.06$\pm$0.38} \\
     \hline
     \multirow{3}*{Cluster-GCN} & Plain & 49.78$\pm$0.25 & 94.32$\pm$0.21 & 80.16$\pm$0.45 & 37.58$\pm$0.33 \\
      & LS & 49.98$\pm$0.41 & 95.18$\pm$0.17  & 80.06$\pm$0.23 & 37.69$\pm$0.31 \\
     & ALS & \textbf{50.20$\pm$0.28} & \textbf{95.24$\pm$0.11} & \textbf{80.78$\pm$0.40} & \textbf{37.89$\pm$0.29} \\
     \hline
      \multirow{3}*{GraphSAINT} & Plain & 51.43$\pm$0.20 & 95.05$\pm$0.13 & 79.07$\pm$0.33  & 47.67$\pm$0.24\\
     & LS & 51.58$\pm$0.23 &  95.02$\pm$0.12 & 79.27$\pm$0.22 & 47.74$\pm$0.38\\
     & ALS & \textbf{51.74$\pm$0.13} & \textbf{95.24$\pm$0.11} & \textbf{79.48$\pm$0.44} & \textbf{47.94$\pm$0.25}\\
     \hline
     \multirow{3}*{MLP} & Plain & 46.36$\pm$0.20 & 70.95$\pm$0.15 & 61.07$\pm$0.21 & \textbf{27.16$\pm$0.18} \\
      & LS & 46.03$\pm$0.93 & 71.61$\pm$0.16 &  61.20$\pm$0.18 & 27.00$\pm$0.22\\
      & ALS & \textbf{46.43$\pm$0.17} & \textbf{71.91$\pm$0.06} & \textbf{61.35$\pm$0.10} & 27.13$\pm$0.28\\
     \hline
    \multirow{3}*{SIGN} & Plain & \textbf{51.15$\pm$ 0.20} & 96.29$\pm$0.03 & 74.04$\pm$0.12 & 19.61$\pm$0.14 \\
      &LS & 51.01$\pm$0.31 & 96.50$\pm$0.04  & 71.07$\pm$0.09 & 18.84$\pm$0.15 \\
     & ALS & 50.84$\pm$0.50 & \textbf{96.56$\pm$0.03} & \textbf{74.20$\pm$0.09} & \textbf{19.77$\pm$0.10} \\
    \bottomrule
  \end{tabular}
  \caption{Test accuracies in percent of the plain model, the model with LS, and the model augmented with ALS. The best performance in each study is in bold.}
  \label{tab: full}
\end{table}

\paragraph{Ablation studies.}
To demonstrate how each module of ALS affects the generalization performance and answer the research question \textbf{Q2}, we perform ablation studies over two sub-graph sampling based backbone networks, i.e., Cluster-GCN and GraphSAINT. In particular, to study the contribution of label propagation in ALS, we ablate it and replace soft label $\bm{y}^{(K)}_i$ with one-hot hard label $\bm{y}_i$ to compute the smoothed label used for model training. The smoothed label is then obtained by: $\bm{y}_i^{ALS} = (1-\alpha_t)\bm{y}_i + \alpha_t\mathrm{Softmax}(\bm{W}\bm{y}_i)$. To ablate the label refinement, we use the soft label $\bm{y}^{(K)}_i$ obtained from the label propagation to compute the smoothed label, i.e., $\bm{y}_i^{ALS} = (1-\alpha_t)\bm{y}_i + \alpha_t \bm{y}^{(K)}_i$. At the same time, we remove the KL distance constraint in Eq.~(\ref{eq: ALS_loss}). 
To ablate smooth pacing module, we use a constant smoothing strength $\alpha$ with $0.1$ to generate smoothed label, i.e., $\bm{y}_i^{ALS} = 0.9*\bm{y}_i + 0.1*\mathrm{Softmax}(\bm{W}\bm{y}^{(K)}_i)$.

We summarize the ablation studies on the three modules of ALS in Table~\ref{tab: ablation}. It is observed that the ablation of any module will decrease the test accuracy, which empirically demonstrate their importances to adapt label smoothing in regularizing the graph representation learning. Comparing with the ablation of label propagation, we observe that the removing of label refinement and smooth pacing extremely damages the performance of ALS. Even with inaccurate prior knowledge of neighborhood label distribution, the label refinement module could be supervised to refine the smoothed label correctly to regularize model prediction. The smooth pacing allocates a smaller smoothing strength $\alpha_t$ at the initial training phase, since the smoothed label is far from being well refined, and then gradually improves  $\alpha_t$ to regularize model from being overfitting. 

\begin{table}[ht]
\setlength{\tabcolsep}{5pt}
  \centering
  \begin{tabular}{c|c|c|c|c|c}
    \toprule
    Backbones & Methods & Flickr & Reddit & ogbn-products & ogbn-mag\\
     \hline
     \hline

     & ALS & \textbf{50.20$\pm$0.28} & \textbf{95.24$\pm$0.11} &  \textbf{80.78$\pm$0.40} & \textbf{37.89$\pm$0.29} \\
     Cluster-& w/o label propagation & 50.06$\pm$0.24 & 95.18$\pm$0.15 & 80.61$\pm$0.40 & 37.85$\pm$0.20\\
     GCN & w/o label refinement & 49.97$\pm$0.25 & 95.06$\pm$0.16 & 80.19$\pm$0.41 & 37.62$\pm$0.28 \\
     & w/o smooth pacing & 50.09$\pm$0.20 & 95.12$\pm$0.13 & 80.53$\pm$0.62 & 37.87$\pm$0.30 \\
     \hline
       & 
     ALS & \textbf{51.74$\pm$0.13} & \textbf{95.24$\pm$0.11} & \textbf{79.48$\pm$0.44} & \textbf{47.94$\pm$0.25}\\
     Graph-& w/o label propagation & 51.59$\pm$0.13 & 95.23$\pm$0.08 & 79.31$\pm$0.44 & 47.91$\pm$0.35 \\
     SAINT& w/o label refinement & 51.67$\pm$0.19 & 95.17$\pm$0.09 & 79.11$\pm$0.55 & 47.79$\pm$0.39\\
     & w/o smooth pacing & 51.64$\pm$0.24 & 95.13$\pm$0.09 & 79.09$\pm$0.61 & 47.83$\pm$0.24\\
    \bottomrule
  \end{tabular}
  \caption{Test accuracies in percent of ALS and its three variants obtained by ablating specific modules.}
  \label{tab: ablation}
\end{table}

\paragraph{Comparison of prior label knowledge.}
Besides the label smoothing, to scale to the large-scale graph, there are two other lines of work to exploit the prior knowledge of neighborhood label distribution $\bm{y}^{(K)}_i$. First, the label propagation method uses $\bm{y}^{(K)}_i$ to predict test nodes without any learnable parameters. Second, one can concatenate $\bm{y}^{(K)}_i$ with node features that are treated as the input for classification model. To answer the research question \textbf{Q3}, we compare these three different methods to exploit $\bm{y}^{(K)}_i$. Specifically, we implement label smoothing and concatenate $\bm{y}^{(K)}_i$ over the GraphSAINT backbone, and directly adopt the proposed label propagation module.  

Table~\ref{tab: label_exploit} summarizes the test accuracies on the four benchmark datasets, where ALS generally achieves the superior performances. The label propagation fails to learn the informatic label prediction on ogbn-mag dataset, since it cannot combine the neighbor labels effectively without modeling the diverse node/edge types in heterogeneous graphs. The label propagation cannot adapt the prior label knowledge to classification tasks without any learnable parameters. Compared with the input concatenation, our label smoothing approach directly regularizes the model prediction to avoid the over-confident prediction and thus obtains better generalization performance. 

\begin{table}[ht]
\setlength{\tabcolsep}{5.5pt}
  \centering
  \begin{tabular}{c|c|c|c|c}
    \toprule
     Methods & Flickr & Reddit & ogbn-products & ogbn-mag \\
     \hline
     \hline
     Label propogation & 50.33 & 92.33 & 73.45 & -\\
     GraphSAINT+Label Input & \textbf{51.75 ± 0.22} & 95.10 ± 0.09 & 76.90 ± 0.32 & 47.20 ± 0.29 \\
     GraphSAINT+ALS & 51.74$\pm$0.13 & \textbf{95.24$\pm$0.11} & \textbf{79.48$\pm$0.44} & \textbf{47.94$\pm$0.25} \\
    \bottomrule
  \end{tabular}
  \caption{Test accuracies in percent of different label exploitation methods.}
  \label{tab: label_exploit}
\end{table}
  
\begin{wraptable}{r}{0.4\textwidth}
\setlength{\tabcolsep}{5.5pt}
  \centering
  \begin{tabular}{c|c|c}
    \toprule
    Backbone & Methods & obgn-products \\
    \hline
    \hline
    \multirow{3}*{C\&S} & Plain &  84.18$\pm$0.07\\
      & LS &  84.22$\pm$0.09  \\
     & ALS &  \textbf{84.28$\pm$0.06} \\
    \bottomrule
  \end{tabular}
  \caption{Test accuracy in percent.}
  \label{tab: CS}
\end{wraptable}

Recently, scalable learning method of C\&S is proposed to refine model prediction in the post-processing step~\cite{huang2020combining}, and shows promising performance on ogbn-products. By training a simple MLP to obtain the initial label predictions, C\&S propagates prediction errors and labels to obtain smoothed prediction results. To demonstrate that ALS is general to any scalable learning framework, we use LS and ALS to regularize the training of MLP module in C\&S, and compare test accuracies in Table~\ref{tab: CS}. By regularizing MLP to obtain  better initial label predictions, our proposed ALS can further improve the test accuracy up to $84.28$, which is the state-of-the-art performance in the leader board. 

\begin{figure}
    \centering
    \includegraphics[width=\textwidth]{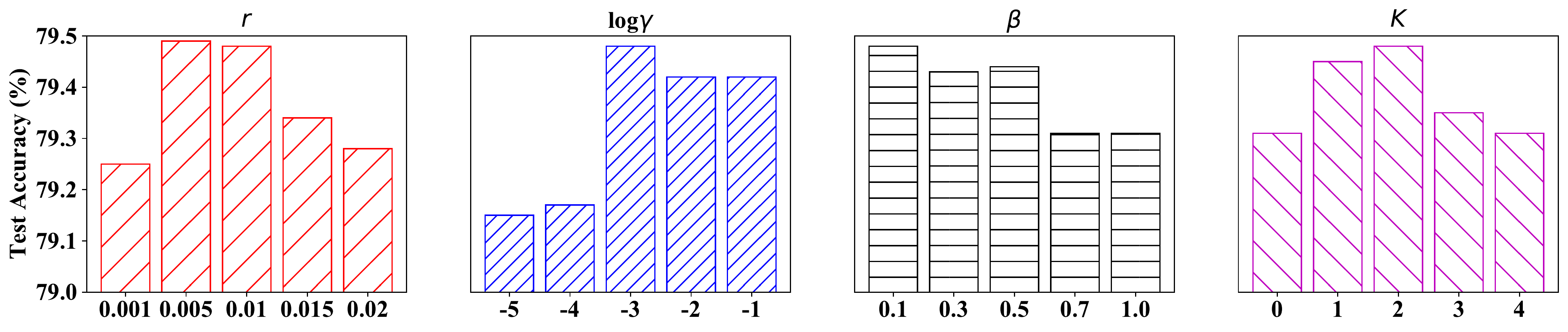}
    \vspace{-0.7cm}
    \caption{Hyperparameter studies of $r$, $\gamma$, $\beta$, and $K$ of GraphSAINT+ALS on ogbn-products.}
    \label{fig: hyper}
\end{figure}

\paragraph{Hyperparameter studies.}
To answer the research question \textbf{Q4}, we conduct experiments with different values of pacing rate $r$, loss hyperparameter $\gamma$, residual strength $\beta$, and label propagation step $K$. Figure~\ref{fig: hyper} illustrates the hyperparameter studies of GraphSAINT+ALS on ogbn-products. In general, within appropriate value ranges, most of these hyperparameter settings can achieve test accuracies larger than $79.3\%$, which is much superior than the baseline GraphSAINT. Specifically, the over-small (or large) pacing rate $r$ damages the test performance due to the insufficient (or excessive) label smoothing regularization. The loss hyperparameter $\gamma$ should be large enough (e.g., $\gamma\geq 10^{-3}$), so as to avoid the learned smoothed label collapsing into one-hot hard target and to guarantee its regularization effect. The superior performance brought by $\beta\leq 0.5$ demonstrates the importance of neighborhood label distribution in learning the structure-aware smoothed label. Based on Eq.~(\ref{eq: lp}), with a smaller $\beta$, we tend to aggregate neighborhood labels during the label propagation. Similar to the common GNN settings, a small value of $K$ is sufficient to aggregate the positively related neighbors to model the smoothed label correctly.

\begin{wrapfigure}{r}{.3\textwidth}
\includegraphics[width=0.3\textwidth]{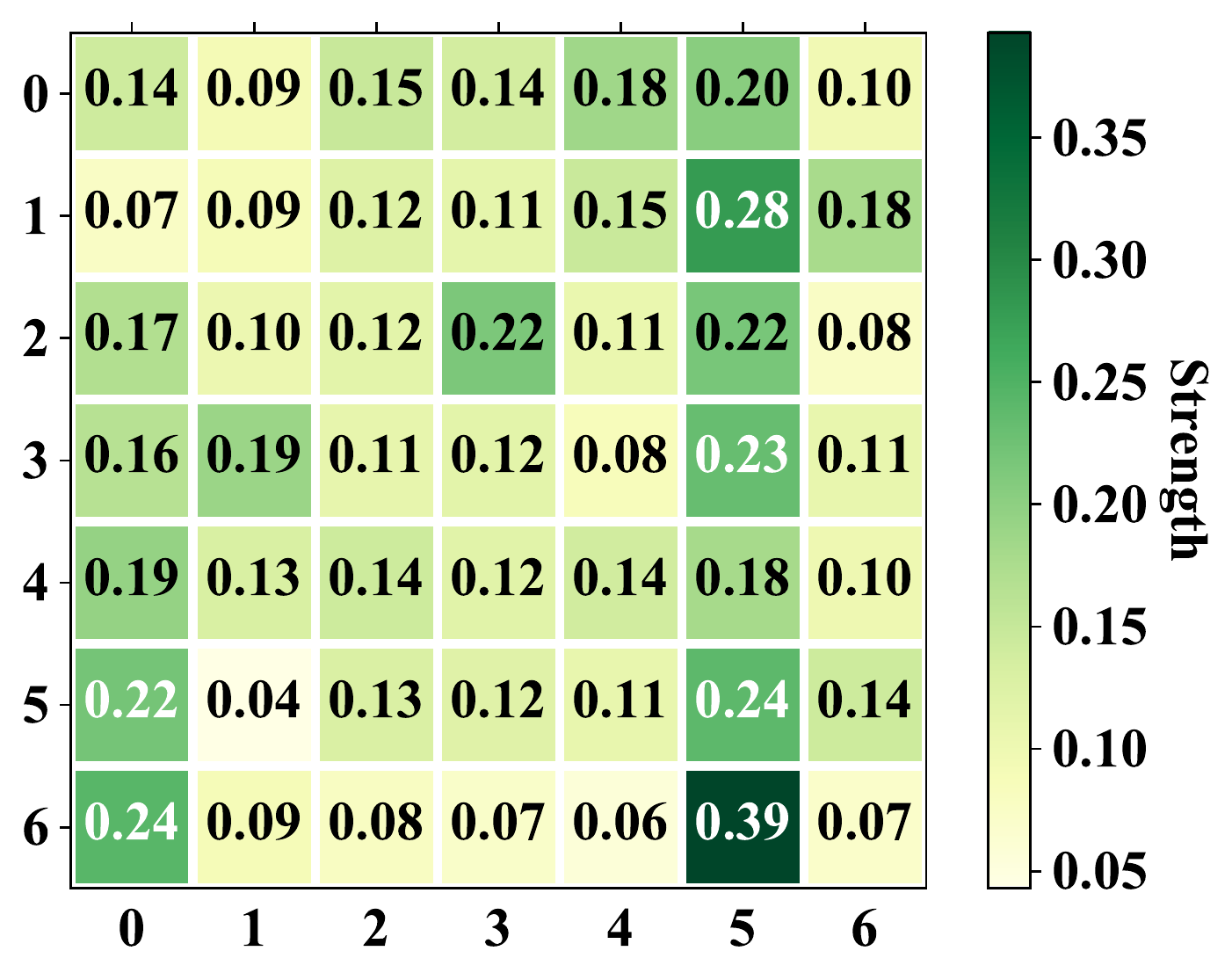}
\caption{Visualization of $\mathrm{Softmax}(\bm{W})$ of ALS on Flickr. X and Y axis denote node label.}
\label{fig:W_strength}
\end{wrapfigure}

\paragraph{Global label relevance visualization.} We visualize $\mathrm{Softmax}$ transformation of the global label relevance matrix along each row, i.e., $\mathrm{Softmax}(\bm{W})$. Note that $\bm{W}$ is learned on backbone framework GraphSAINT and dataset Flickr. As shown in Figure~\ref{fig:W_strength}, we observe that each class label has unbalanced relevance strengths to the other classes. This is in line with our motivation that the pair-wise label relevance is different from each other, and far away from the uniform distribution. As demonstrated in the previous experiments, the modeling of such global label relevance  delivers the superior performances comparing with LS.

\section{Related Work}

\paragraph{Graph neural networks.} GNNs have shown superiority in processing graphs, i.e., data with rich relational structures. GNNs could be categorized into spectral domain and spatial domain models. The spectral models~\cite{bruna2013spectral} extends convolution on images to graph data by modeling on the spectrum of graph Laplacian. Models designed from the spatial perspective simplify the spectral models. Spatial models such as ChebNet~\cite{defferrard2016convolutional}, GCN~\cite{kipf2016semi}, GAT~\cite{velivckovic2017graph} and GIN~\cite{xu2018powerful}, could also be understood from the message passing perspective. GNNs are playing increasingly crucial roles in various applications such as recommender systems~\cite{ying2018graph, chen2020revisiting, zhou2021temporal}, social network analysis~\cite{fan2019graph, zhou2019auto, zhou2020towards, zhou2021dirichlet}, and biochemical module analysis~\cite{gilmer2017neural, zhou2019multi}. 

\paragraph{Scalable graph representation learning.}  Two types of methods have been proposed to tackle scalability issue of GNNs, including sub-graph sampling methods~\cite{hamilton2017inductive, ying2018graph, chen2018fastgcn, gao2018large, chen2017stochastic, chiang2019cluster, zeng2019graphsaint} and precomputing methods~\cite{wu2019simplifying, rossi2020sign, bojchevski2020scaling}. 
To reduce computation and memory cost, the sub-graph sampling methods feed GNNs only with a small batch of sub-graphs, which consist of subsets of nodes and edges.  Specifically, 
ClusterGCN~\cite{chiang2019cluster} conducts training on sampled sub-graphs in each batch, but the sub-graphs are obtained through clustering algorithms. GraphSAINT~\cite{zeng2019graphsaint} samples sub-graphs that are appropriately connected for information propagation, where a normalization technique is also proposed to eliminate bias. The major limitation for sub-graph based methods is that distant nodes in the original graph are unlikely to be fed into the GNNs in the same batch, thus leading to label bias in the trained models. The precomputing methods of SIGN~\cite{rossi2020sign} and SGC\cite{wu2019simplifying} remove trainable weights, and propagate node features over the graph in advance to store the smoothed features. 


\paragraph{Label propagation.} Label propagation distributes the observed node labels over the graph following the connection between nodes~\cite{zhou2004learning, zhu2005semi}. It has been used for semi-supervised training on graph data~\cite{zhang2006hyperparameter}, where node labels are partially observed. The assumption behind is that labels and features change smoothly over the edges of the graph. It has also been proposed to combine feature propagation with label propagation towards a unified message passing scheme~\cite{shi2020masked}. Some recent work connects GNNs with label propagation~\cite{wang2020unifying, jia2020residual} by studying how labels/features spread over a graph and how the initial feature/label of one node influences the prediction of another node.

\paragraph{Label smoothing.} Label smoothing improves the generalization~\cite{szegedy2016rethinking, pereyra2017regularizing} and robustness~\cite{pang2017towards, goibert2019adversarial, shen2019defending} of a deep neural network. Label smoothing replaces one-hot labels with smoothed labels. 
It has been shown that label smoothing has similar effect as randomly replacing some of the ground-truth labels with incorrect values at each mini-batch~\cite{xie2016disturblabel}. \cite{pang2017towards} proposes reverse cross entropy for gradient smoothing. It encourages a model to better distinguish adversarial examples from normal ones in representation space. ~\cite{wang2020inference} proposes the graduated label smoothing method, where high-confidence predictions are assigned with higher smoothing penalty than low-confidence ones. 

\section{Conclusions}
In this paper, we point out the inherent label bias within the sampled sub-graphs for the mini-batch training of large-scale graph. By minimizing vanilla cross-entropy loss, we empirically analyze that such label bias will make GNN model over-confidently predict the ground-truth class and lead to overfitting issue.  To overcome the label bias and the resulted over-confident prediction, we propose an adaptive label smoothing to replace the one-hot hard target with smoothed label, which allocates prediction confidence to other classes to avoid overfitting. Specially, we learn the smoothed label with the prior knowledge of local neighborhood label distribution and the global label refinement to adapt to graph data on hand. The experiments show that our algorithm could generally improve the test performance by relieving the overfitting on biased labels.

\bibliographystyle{abbrv}
\bibliography{ref}

\newpage
\appendix

\section{Appendix}
\subsection{Datasets}
The dataset statistics of Flickr~\cite{zeng2019graphsaint}, Reddit~\cite{hamilton2017inductive}, obgn-products~\cite{hu2020open}, and ogbn-mag~\cite{hu2020open} are listed in Table~\ref{tab:dataset}. \textit{Flickr} is a social network, where the nodes represent images and the edges denote the shared properties between two images. The node classification task in Flickr is to categorize the types of images. \textit{Reddit} is a social netowork, where the nodes  represent posts in Reddit
forum and the edges indicate the same user comments between two posts. The node classification task in Reddit is to predict the communities of online posts based on user comments. \textit{ogbn-products} is an Amazon product co-purchasing network, where the nodes represent products sold in Amazon and the edges  indicate the co-purchasing relationships between two products. The node classification task in ogbn-products is to predict the category of a product. \textit{ogbn-mag} is a heterogeneous network extracted from the Microsoft Academic Graph. It contains four types of entities: papers, authors, institutions, and fields of study. The directed edges are categorized into four types--an author is “affiliated with” an institution, an author “writes” a paper, a paper “cites” a paper, and a paper “has a topic of” a field of study. The node classification task in ogbn-mag is to predict the venue (conference or journal) of each entity of paper.

\begin{table*}[h]
\setlength{\tabcolsep}{3pt}
    \centering
    \begin{tabular}{c|ccccc}
    \toprule
      Datasets  & \# Nodes & \# Edges & \# Classes & \# Features & \# Train/Validation/Test \\
      \hline
      Flickr  &  89,250 & 899,756 & 7 & 500 &  0.50 / 0.25 / 0.25 \\
      Reddit &  232,965 & 11,606,919 & 41 & 602 &   0.66 / 0.10 / 0.24\\
      ogbn-products & 2,449,029 &  61,859,140 & 47 & 100 & 0.08 / 0.02 / 0.90 \\
      ogbn-mag &  1,939,743 &  21,111,007 & 349 &128 & 0.85 / 0.09 / 0.06\\
    \bottomrule  
    \end{tabular}
    \caption{Data statistics.}
    \label{tab:dataset}
\end{table*}

\subsection{Backbone Frameworks}
We evaluate our ALS on two main categories of scalable graph representation learning frameworks: one is based on the sub-graph sampling and the other one is based on precomputing. Although we aim at solving the label bias and over-confident prediction in the sub-graph sampling methods, we show that our method is general to both of these two scalable backbone frameworks. Specifically, we adopt the following five representative backbone frameworks:
\begin{itemize}[leftmargin=*, topsep=0pt, noitemsep]
    \item \textit{GraphSAGE}~\cite{hamilton2017inductive} (sub-graph sampling based). It is a node-wise sampling method to uniformly sample a batch of training nodes and their neighbors of different orders. The sampled nodes and neighbors construct several sub-graphs to formulate a batch. 
    \item \textit{Cluster-GCN}~\cite{chiang2019cluster} (sub-graph sampling based). It first conducts node clustering algorithm to partition the input graph into a series of sub-graphs. During the training phase, each batch is directly formulated by a random subset of preprocessed sub-graphs.
    \item \textit{GraphSAINT}~\cite{zeng2019graphsaint} (sub-graph sampling based). Starting from a subset of training nodes, we choose the random walk sampler (i.e., the most powerful one as reported in~\cite{zeng2019graphsaint}) to sample their neighbors for constructing sub-graphs in one batch.
    \item \textit{MLP} (precomputing based). MLP is widely used to classify nodes based on the precomputed node features. Herein, MLP directly uses the original node features, which has been shown to achieve good classification performance in the graph data. In the precomputing methods, each node can be regarded as an independent sample, and does not connect to its neighbors. The batch is thus directly represented by an independent subset of training nodes. 
    \item \textit{SIGN}~\cite{rossi2020sign} (precomputing based). In the preprocessing step, SIGN conducts message-passing strategy and precomputes node features as: $\hat{\bm{A}}^{l}X$ for $l\in \{1, \cdots, L\}$. $\hat{\bm{A}}$ is a normalized adjacency matrix used in GCN~\cite{kipf2016semi}. The precomputed node features of different orders are concatenated together to augment the original node features. In the training phase, the batch is constructed by a random subset of training nodes, and taken as input to the downstream classification model of MLP.
\end{itemize} 

We implement the above scalable backbone frameworks according to the official examples of Pytorch Geometric\footnote{https://github.com/rusty1s/pytorch\_geometric}. The basic model hyperparameters are defined in the examples or determined according to their public literature, including batch size, learning rate, weight decay, training epochs, hidden units, dropout rate, etc. All of the sub-graph sampling methods apply a three-layer GNN model, while the precomputing methods use a three-layer MLP. Following the official examples, we use full batch training in the precomputing methods.

\subsection{Implementation Details}
We further implement LS and ALS over each backbone model.  For LS with uniform distribution, we set the constant smoothing strength $\alpha$ as $0.1$, which is widely applied in regularizing image classification. For our ALS, we use the linear pacing and exponential pacing functions for the sub-graph sampling methods and precomputing methods, respectively. To have a fair comparison with LS, we set $\alpha_{\max}=0.1$ in our ALS. For each combination of backbone framework and dataset, we choose the appropriate hyperperameters of residual strength $\beta$ and step $K$ in the label propagation, and also determine the KL distance constraint $\gamma$ as well as the smooth pacing rate $r$. The detailed hyperparameters involved in ALS are shown in Table~\ref{tab: hyper}. Notably, comparing with the sub-graph sampling methods, we use the negative pacing rate $r$ in the precomputing methods. Instead of using mini-batch training, the official examples of MLP and SIGN applies the full batch training. That means the trainable parametere $W$ and soft label $\bm{y}_i^{\mathrm{soft}}$ could be well updated at the initial training phase. Therefore, we use the decreasing smoothing strength in the precomputing methods, where the models are regularized strictly by the difficult smoothed label from the beginning. At the end of training, the precomputing methods are relaxed to learn the easy one-hot hard target to improve the test performance. The initial smoothing strengths $b$ in the exponential pacing function for MLP are: $0.05$ in Flickr, $0.2$ in Reddit, and $0.08$ in ogbn-products \& ogbn-mag. The values of $b$ for SIGN are: $0.1$ in Flickr, $0.2$ in Reddit, $0.15$ in ogbn-products, and $0.08$ in ogbn-mag.

\begin{table}[t]
\setlength{\tabcolsep}{3.pt}
  \centering
  \begin{tabular}{c|cccc|cccc|cccc|cccc}
    \toprule
    \multirow{2}*{Backbones} & \multicolumn{4}{c|}{Flickr} & \multicolumn{4}{c|}{Reddit} & \multicolumn{4}{c|}{ogbn-products} & \multicolumn{4}{c}{ogbn-mag} \\
    \cline{2-17}
    & $r$ & $\gamma$ & $\beta$ & $K$ & $r$ & $\gamma$ & $\beta$ & $K$ & $r$ & $\gamma$ & $\beta$ & $K$ & $r$ & $\gamma$ & $\beta$ & $K$ \\
     \hline
     \hline
     GraphSAGE & 5e-3 & 5e-3 & 0.5 & 2 & 2e-2 & 1e-2 & 0 & 5 & 1e-2 & 1e-3 & 0.1 & 5 & 1e-2 & 1e-6 & 1e-2 & 5 \\
     Cluster-GCN & 1e-2 & 1e-2 & 0.1 & 2 & 2e-2 & 1e-2 & 0 & 5 & 5e-3 & 1e-2 & 0 & 5 & 0.1 & 1e-6 & 1e-2 & 5 \\
     GraphSAINT & 5e-3 & 5e-3 & 0.5 & 2 & 2e-3 & 1e-3 & 0 & 5 & 1e-2 & 1e-3 & 0.1 & 2 & 2e-3 & 1e-3 & 0.1 & 2 \\
     MLP &  -2e-2 & 1e-3 & 0.5 & 2 & -1e-3 & 1e-2 & 0 & 5 & -1e-2 & 1e-2 & 0.1 & 10 & -1e-2 & 1e-3 & 0.8 & 2\\
     SIGN & -0.2 & 1e-3 & 0.1 & 2 & -5e-3 & 1e-3 & 0 & 5 & -0.1 & 1e-6 & 0.1 & 5 & -2e-2 & 1e-3 & 0.8 & 2 \\
    \bottomrule  
  \end{tabular}
  \caption{Hyperparameter choices in ALS for each combination study of backbone and dataset.}
  \label{tab: hyper}
\end{table}

\subsection{Running Environment}
All the experiments are implemented with PyTorch, and tested on a machine with 24 Intel(R) Xeon(R) CPU E5-2650 v4 @ 2.20GB processors, 128GB CPU memory size, and one GPU of  GeForce RTX 3090 with 24 GB memory size.

\subsection{Label Bias, Over-confident Prediction and Overfitting Observations}
\begin{figure}[htb]
    \centering
    \includegraphics[width=\textwidth]{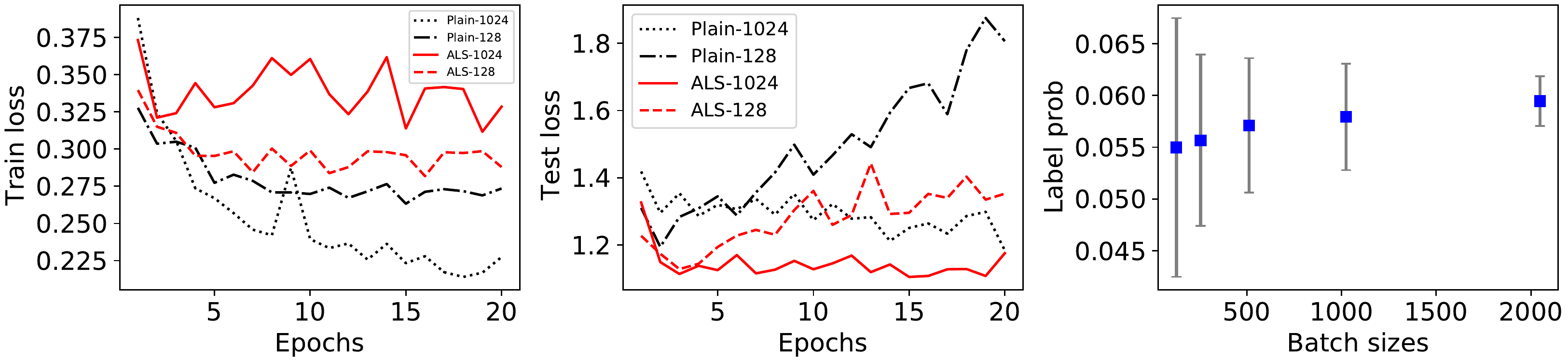}
    \caption{\textbf{Left, Middle:} The training and testing losses upon backbone GraphSAGE and dataset ogbn-products. While Plain-$a$ means the original backbone trained with batch size of $a$, ALS-$a$ denotes one equipped with ALS. \textbf{Right:} The mean probability $p_c$ of nodes with specific class $c$ within a batch (i.e., blue square), and the standard variance of probability $p_c$ among batches (i.e., upper and lower bar). Herein we show $c=0$ on  ogbn-products for an example.}
    \label{fig:GraphSAGE_products}
\end{figure}

\begin{figure}[htb]
    \centering
    \includegraphics[width=\textwidth]{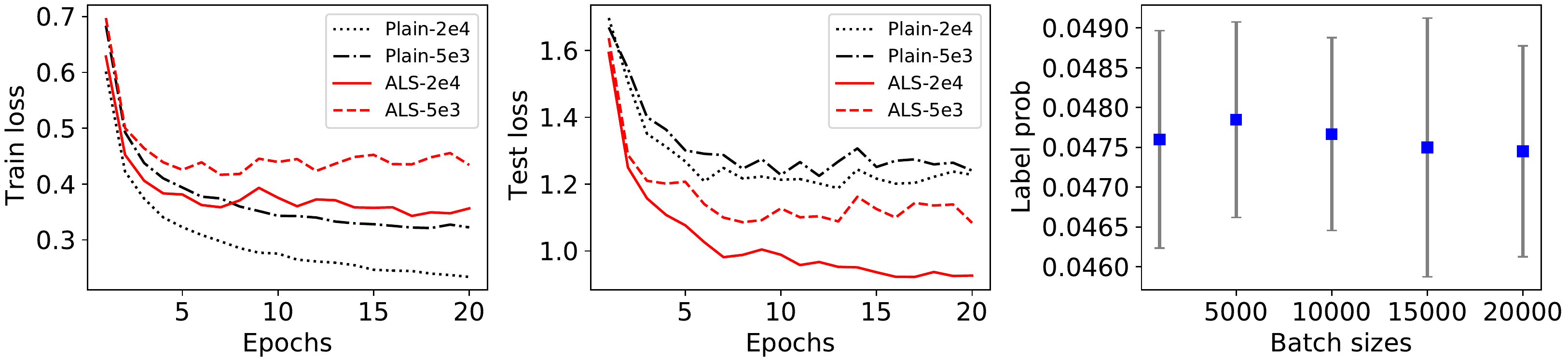}
    \caption{\textbf{Left, Middle:} The training and testing losses upon backbone GraphSAINT and dataset ogbn-products. While Plain-$a$ means the original backbone trained with batch size of $a$, ALS-$a$ denotes one equipped with ALS. \textbf{Right:} The mean probability $p_c$ of nodes with specific class $c$ within a batch (i.e., blue square), and the standard variance of probability $p_c$ among batches (i.e., upper and lower bar). Herein we show $c=0$ on  ogbn-products for an example.}
    \label{fig:GraphSAINT_products}
\end{figure}

\begin{figure}[htb]
    \centering
    \includegraphics[width=\textwidth]{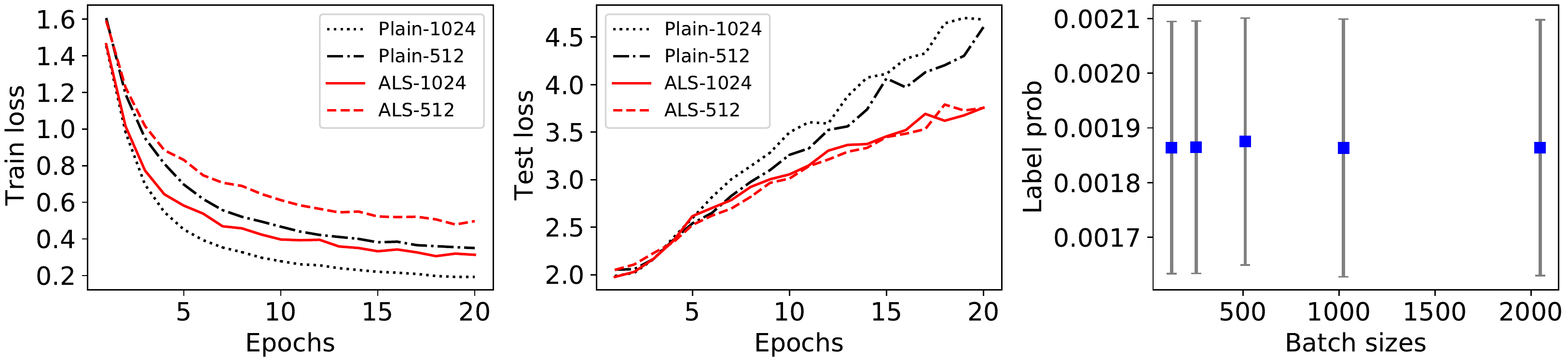}
    \caption{\textbf{Left, Middle:} The training and testing losses upon backbone GraphSAGE and dataset ogbn-mag. While Plain-$a$ means the original backbone trained with batch size of $a$, ALS-$a$ denotes one equipped with ALS. \textbf{Right:} The mean probability $p_c$ of nodes with specific class $c$ within a batch (i.e., blue square), and the standard variance of probability $p_c$ among batches (i.e., upper and lower bar). Herein we show $c=0$ on  ogbn-mag for an example.}
    \label{fig:GraphSAGE_mag}
\end{figure}

\begin{figure}[htb]
    \centering
    \includegraphics[width=\textwidth]{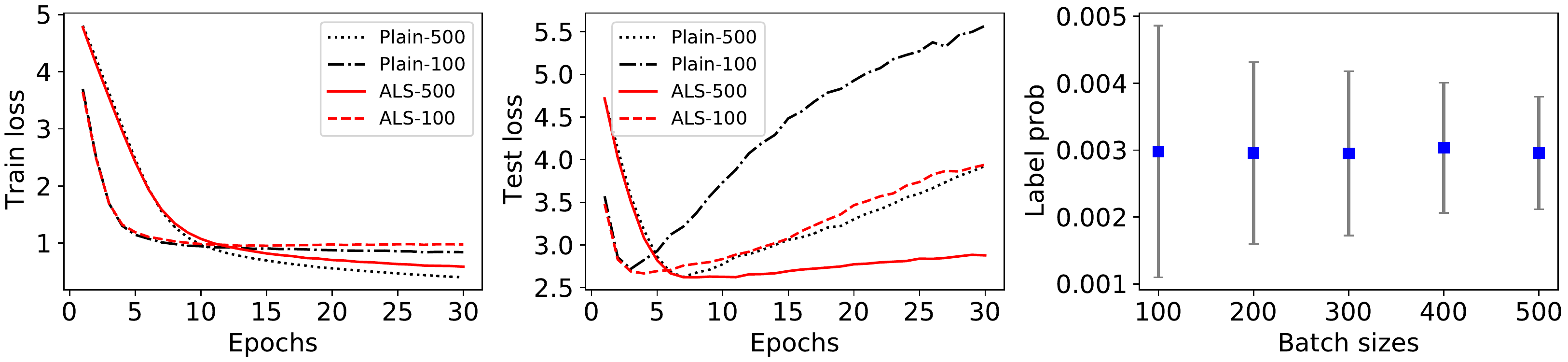}
    \caption{\textbf{Left, Middle:} The training and testing losses upon backbone Cluster-GCN and dataset ogbn-mag. While Plain-$a$ means the original backbone trained with batch size of $a$, ALS-$a$ denotes one equipped with ALS. \textbf{Right:} The mean probability $p_c$ of nodes with specific class $c$ within a batch (i.e., blue square), and the standard variance of probability $p_c$ among batches (i.e., upper and lower bar). Herein we show $c=0$ on  ogbn-mag for an example.}
    \label{fig:Cluster_GCN_mag}
\end{figure}

\begin{figure}[htb]
    \centering
    \includegraphics[width=\textwidth]{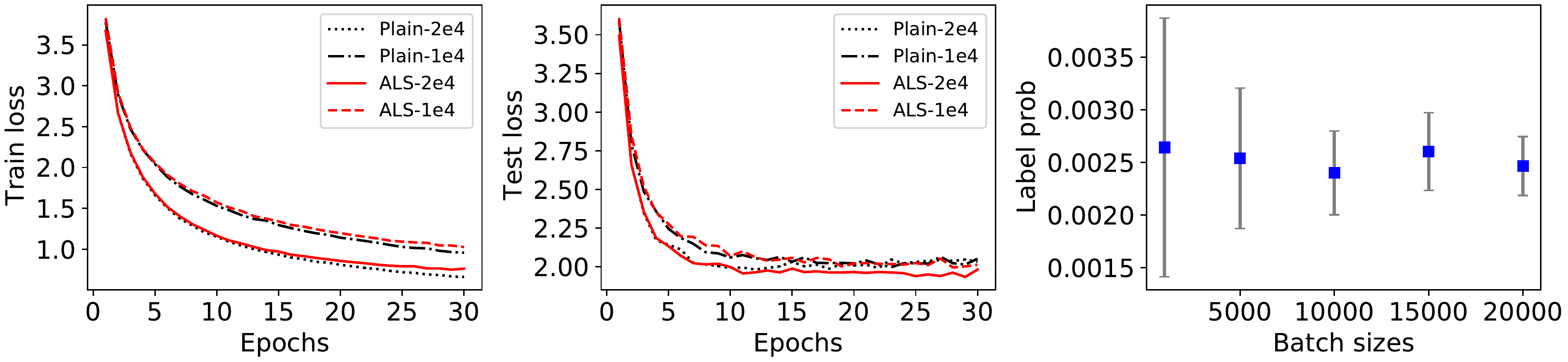}
    \caption{\textbf{Left, Middle:} The training and testing losses upon backbone GraphSAINT and dataset ogbn-mag. While Plain-$a$ means the original backbone trained with batch size of $a$, ALS-$a$ denotes one equipped with ALS. \textbf{Right:} The mean probability $p_c$ of nodes with specific class $c$ within a batch (i.e., blue square), and the standard variance of probability $p_c$ among batches (i.e., upper and lower bar). Herein we show $c=0$ on  ogbn-mag for an example.}
    \label{fig:GraphSAINT_mag}
\end{figure}

\begin{figure}[htb]
    \centering
    \includegraphics[width=\textwidth]{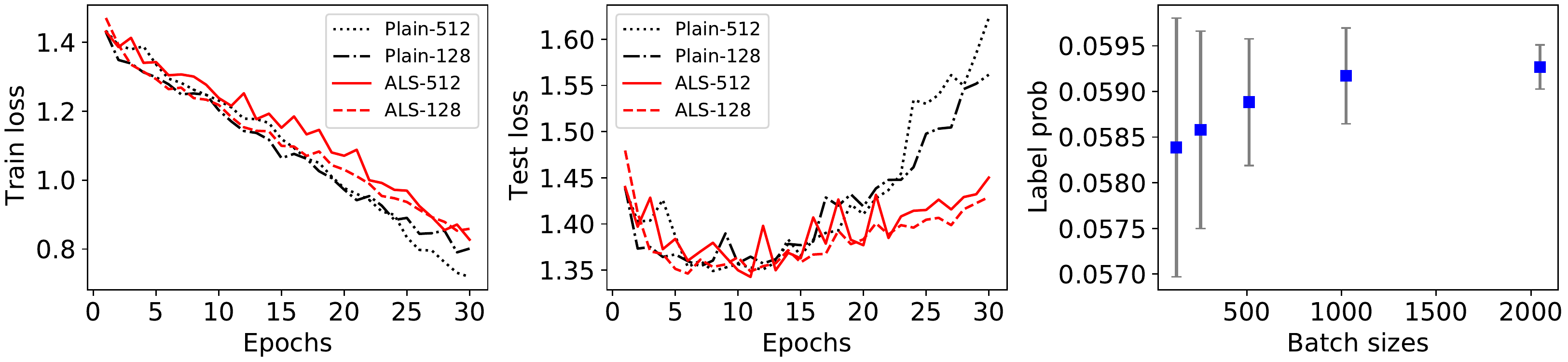}
    \caption{\textbf{Left, Middle:} The training and testing losses upon backbone GraphSAGE and dataset Flickr. While Plain-$a$ means the original backbone trained with batch size of $a$, ALS-$a$ denotes one equipped with ALS. \textbf{Right:} The mean probability $p_c$ of nodes with specific class $c$ within a batch (i.e., blue square), and the standard variance of probability $p_c$ among batches (i.e., upper and lower bar). Herein we show $c=0$ on  Flickr for an example.}
    \label{fig:GraphSAGE_Flickr}
\end{figure}

\begin{figure}[htb]
    \centering
    \includegraphics[width=\textwidth]{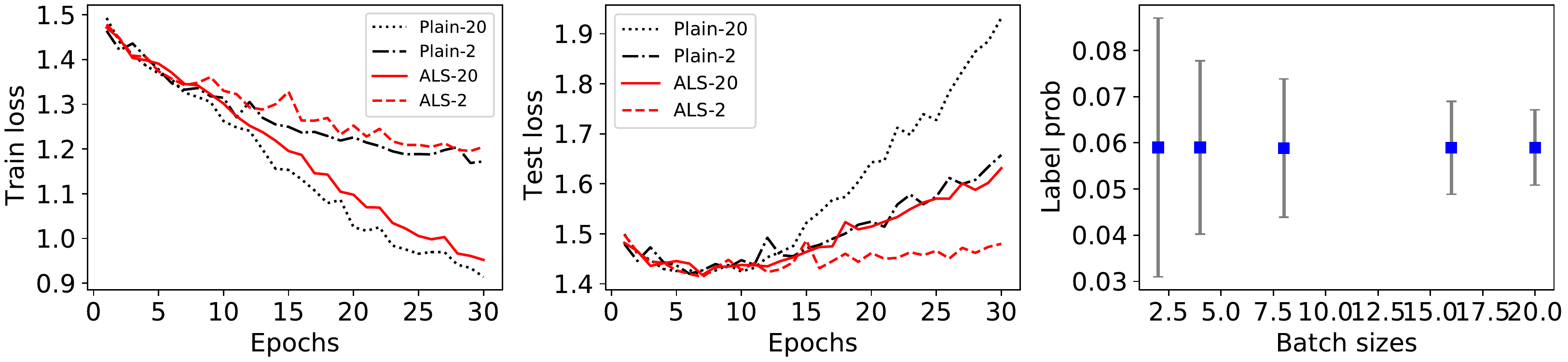}
    \caption{\textbf{Left, Middle:} The training and testing losses upon backbone Cluster-GCN and dataset Flickr. While Plain-$a$ means the original backbone trained with batch size of $a$, ALS-$a$ denotes one equipped with ALS. \textbf{Right:} The mean probability $p_c$ of nodes with specific class $c$ within a batch (i.e., blue square), and the standard variance of probability $p_c$ among batches (i.e., upper and lower bar). Herein we show $c=0$ on  Flickr for an example.}
    \label{fig:Cluster_GCN_Flickr}
\end{figure}

\begin{figure}
    \centering
    \includegraphics[width=\textwidth]{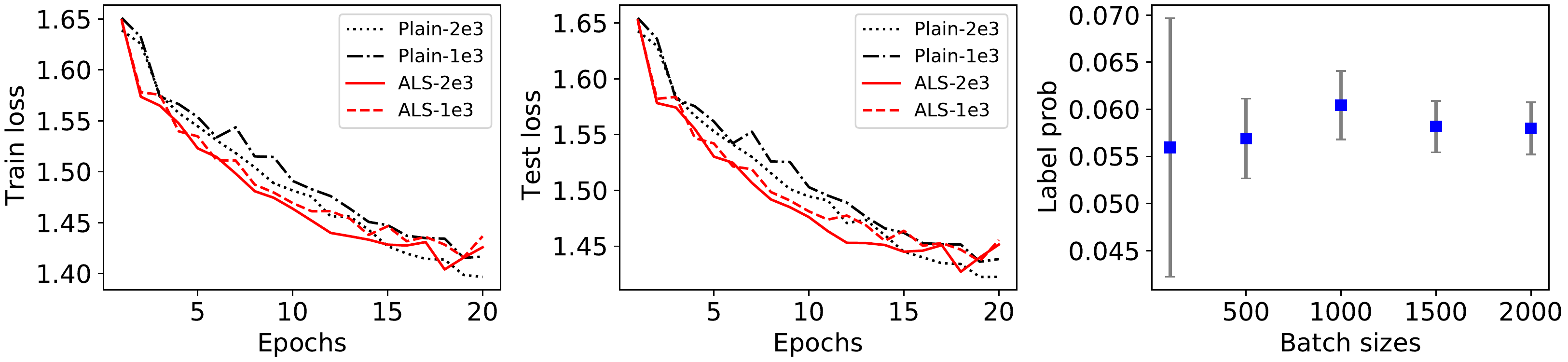}
    \caption{\textbf{Left, Middle:} The training and testing losses upon backbone GraphSAINT and dataset Flickr. While Plain-$a$ means the original backbone trained with batch size of $a$, ALS-$a$ denotes one equipped with ALS. \textbf{Right:} The mean probability $p_c$ of nodes with specific class $c$ within a batch (i.e., blue square), and the standard variance of probability $p_c$ among batches (i.e., upper and lower bar). Herein we show $c=0$ on  Flickr for an example.}
    \label{fig:last}
\end{figure}

As shown in Figure~\ref{fig:loss}, the sub-graph sampling method of Cluster-GCN brings label bias, and leads to the over-confident prediction and overfitting in training set. These problems will damage the model's generalization performance in testing set. Our ALS learns to replace the one-hot hard target $\bm{y}_i$ with the smoothed label $\bm{y}_i^{ALS}$, which could relieve these three problems to improve the generalization ability. In this section, we report the training losses, the testing losses, and the label biases of all the sub-graph sampling methods on ogbn-products, ogbn-mag, and Flickr. We show the experimental results in Figures~\ref{fig:GraphSAGE_products}-\ref{fig:last}. Note that the batch sizes of GraphSAGE, Cluster-GCN, and GraphSAINT are defined by the corresponding sub-graph sampling functions in Pytorch Geometric\footnote{https://pytorch-geometric.readthedocs.io/en/latest/modules/data.html}, i.e., NeighborSampler, ClusterLoader, and GraphSAINTRandomWalkSampler. While the batch sizes of GraphSAGE and GraphSAINT specify how many training samples per batch to load, the batch size of Cluster-GCN determines how many clustered sub-graphs to  sample. We make the following empirical observations:
\begin{itemize}
    \item All the sub-graph sampling methods bring label bias within a batch. It is shown that the standard deviance of $p_c$ is extremely large, which is compatible with the mean value of $p_c$. In other word, the nodes within a small batch tend to belong to certain classes, and the label distributions vary dramatically between batches. In general, the smaller the batch size is, the larger the standard deviance of $p_c$ will be. 
    \item Comparing with the plain backbone frameworks, our ALS has larger training losses. That is because ALS replaces the one-hot hard target $\bm{y}_i$ with the smoothed label $\bm{y}_i^{ALS}$, which distributes label confidences to both ground-truth class and the other classes. By minimizing the regularized loss in Eq.~(\ref{eq: ALS_loss}), ALS reduces the model's prediction probability on the ground-truth class, and thus increases the training loss. The regularized prediction probability on the ground-truth class will help the model avoid the over-confident prediction and the overfitting on the training set. 
    \item Comparing with the plain backbone frameworks, our ALS generally has smaller testing losses and better generalization ability. Since the model is regularized to avoid the over-confident prediction, the smooth prediction probability in ALS is more easier to be generalized to the unseen testing set.

\end{itemize}

\end{document}